\documentclass[reprint,aps,pre]{revtex4-1}

\usepackage{amsmath,amsfonts,amscd,amssymb,graphicx,amsthm}

\theoremstyle{definition}

\usepackage{subeqnar}
\usepackage{dcolumn}
\usepackage{booktabs}
\usepackage{overpic}
\usepackage{color}
\usepackage[caption=false]{subfig}
\usepackage{adjustbox}
\usepackage{rotating}
\usepackage{float}
\makeatletter
\let\newfloat\newfloat@ltx
\makeatother
\usepackage{algorithm}
\usepackage{algpseudocode}
\algnewcommand\algorithmicinput{\textbf{Input:}}
\algnewcommand\Input{\item[\algorithmicinput]}

\begin{document}

\title{Accelerating scientific discovery with the {\em common task framework}}

\author{J. Nathan Kutz$^{a,*}$, Peter Battaglia$^b$, 
Michael Brenner$^c$,
Kevin Carlberg$^d$, 
Aric Hagberg$^e$,
Shirley Ho$^f$,
Stephan Hoyer$^g$, 
Henning Lange$^h$, 
Hod Lipson$^{i,*}$,
Michael W. Mahoney$^j$,
Frank Noe$^k$,
Max Welling$^m$, 
Laure Zanna$^n$,
Francis Zhu$^{o,*}$,
Steven L. Brunton$^{p,*}$}

\affiliation{$^*$ AI Institute in Dynamic Systems, Seattle, WA 98195}
\affiliation{$^a$ Department of Applied Mathematics and Electrical and Computer Engineering, University of Washington, Seattle, WA, 98195}
\affiliation{$^b$ Google DeepMind, London, UK}
\affiliation{$^c$ School of Engineering and Applied Physics, Harvard University, Cambridge MA 02138}
\affiliation{$^d$ Meta Platforms, Inc. 1 Hacker Way, Menlo Park, CA, 94025}
\affiliation{$^e$ Computer, Computational, and Statistical Sciences Division, Los Alamos National Laboratory, Los Alamos NM}
\affiliation{$^f$ Center for Computational Astrophysics, Flatiron Institute, 162 5th Avenue, New York, NY 10010}
\affiliation{$^g$ Google Research, Mountain View, California 94043, USA}
\affiliation{$^h$ Amazon Research, Seattle, WA}
\affiliation{$^i$ Department of Mechanical Engineering, Columbia University, New York, NY}
\affiliation{$^j$ Department of Statistics, University of California at Berkeley, International Computer Science Institute, and Lawrence Berkeley National Laboratory, Berkeley, CA}
\affiliation{$^k$ Freie Universit\"{a}t Berlin, Department of Physics, Arnimallee 6, 14195 Berlin, Germany and AI4Science, Microsoft Research, Karl-Liebknecht Str. 32, Berlin, 10178, Germany}
\affiliation{$^m$ University of Amsterdam, Informatics Institute, 94323, 1090 GH  Amsterdam}
\affiliation{$^n$ Courant Institute of Mathematical Sciences, New York University, New York, NY}
\affiliation{$^o$ Hawai‘i Institute of Geophysics and Planetology
University of Hawai‘i at Mānoa, Honolulu, HI 96822}
\affiliation{$^p$ Department of Mechanical Engineering, University of Washington, Seattle, WA 98195}


\begin{abstract}
Machine learning (ML) and artificial intelligence (AI) algorithms are transforming and empowering the characterization and control of dynamic systems in the engineering,  physical, and biological sciences.  These emerging modeling paradigms require comparative metrics to evaluate a diverse set of scientific objectives, including forecasting, state reconstruction, generalization, and control, while also considering limited data scenarios and noisy measurements.   We introduce a common task framework (CTF) for science and engineering, which features a growing collection of challenge data sets with a diverse set of practical and common objectives.  The CTF is a critically enabling technology that has contributed to the rapid advance of ML/AI algorithms in traditional applications such as speech recognition, language processing, and computer vision.  
There is a critical need for the objective metrics of a CTF to compare the diverse algorithms being rapidly developed and deployed in practice today across science and engineering. 
\end{abstract}
\maketitle





Data-science, especially {\em machine learning} (ML) and {\em artificial intelligence} (AI), is transforming almost every aspect of the engineering, physical, social, and biological sciences.   This transformation is driven by the confluence of a number of emerging technologies, including computational hardware (storage and computing), optimization algorithms, open source software, and data collection (sensors).   Powered by Moore's law, the capabilities of modern scientific computing architectures has engendered new scientific exploration paradigms that are readily accessible to a broad range or practitioners.   Within the ML/AI communities, the growing diversity of computational solutions has necessitated the {\em common task framework} (CTF) to provide a critical role in evaluating methodological advancements.  Donoho~\cite{donoho201750} has argued that the successful application of CTFs is a primary factor for the success of data science and machine learning.  Indeed, the fields of {\em speech recognition}, {\em natural language processing}, and {\em computer vision} have developed mature CTF platforms that are progressively updated with more challenging data in order to drive progress and innovation.  For instance, the industry-leading {\em computer vision and pattern recognition} (CVPR) conference offers more than 30 challenge problems per year for participants to score and benchmark their ML/AI algorithms against. More broadly, classic fields of machine learning have benefited from extensive benchmark environments and common task frameworks, including ImageNet~\cite{deng2009imagenet,Krizhevsky2012nips}, Go and chess~\cite{silver2018general}, video games such as Atari~\cite{mnih2015human} and Starcraft~\cite{vinyals2019grandmaster}, the OpenAI Gym~\cite{ravichandiran2018hands,dutta2018reinforcement}, among other environments for more realistic control~\cite{deisenroth2011pilco,todorov2012mujoco}.
While these fields in particular have adopted the CTF broadly, many scientific disciplines have yet to integrate the CTF into their core infrastructure~\cite{mcgreivy2024weak}.
This compromises the true comparative metrics between methods, algorithms, and results, and it limits the potential of ML in these areas.  

The CTF for science and engineering is primarily focused on evaluating machine learning and AI models for dynamic systems, or those systems whose underlying evolution is determined by physical or biophysical principles or governing equations. The CTF will provide training data sets with clear and concise goals related to forecasting and reconstruction under various challenging scenarios, such as noisy measurements, limited data, or varying system parameters.  The user is required to produce approximations for hidden test data. For ease of use, the training and test sets are simple NumPy arrays~\cite{harris2020array}.  The approximations to the test set are evaluated and scored by a referee, with the diverse metrics evaluated and scores posted on a leaderboard.  

\begin{figure*}
    \centering
    \includegraphics[width=.8\textwidth]{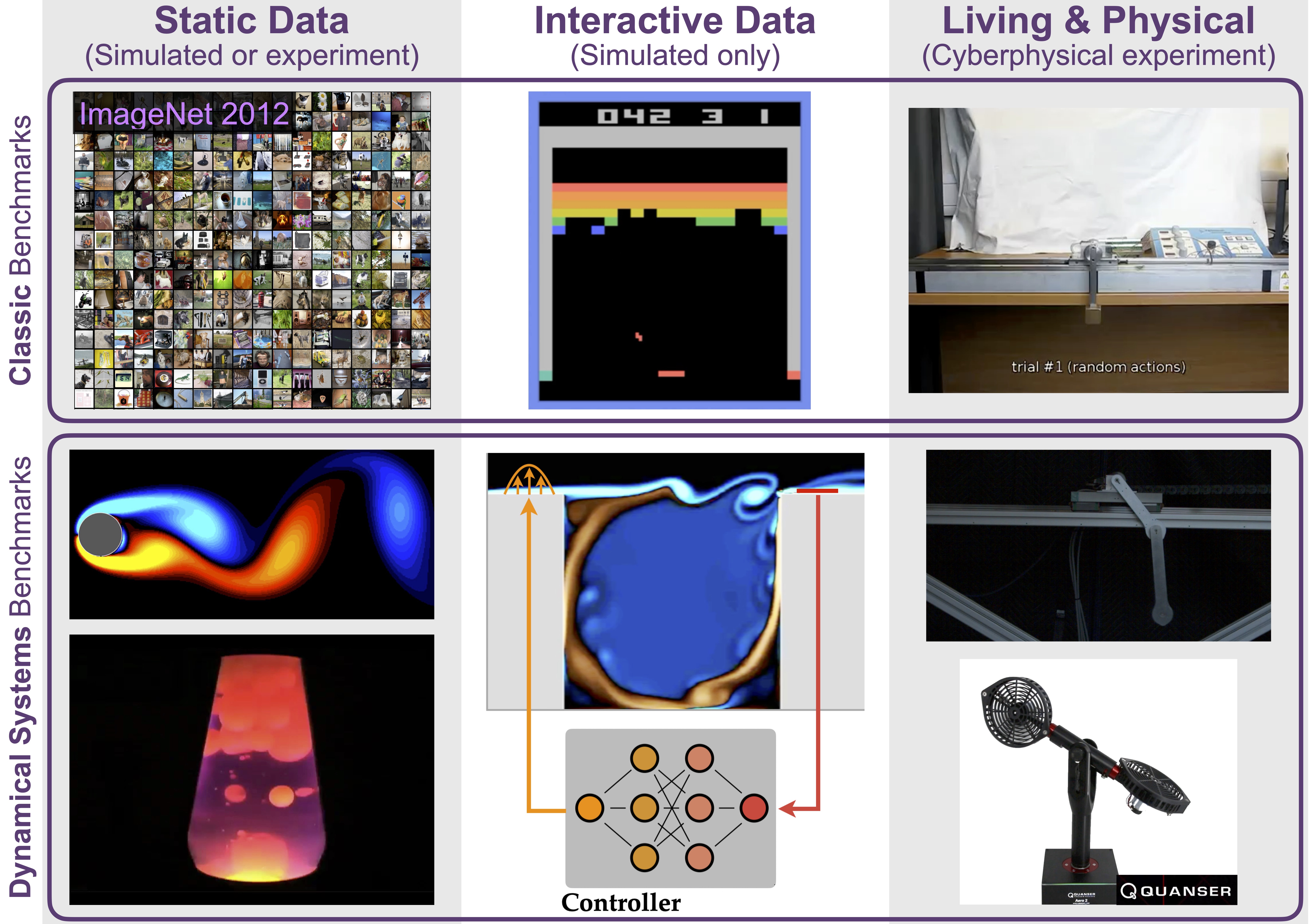}
    \caption{Illustration of classic CTF benchmark problems and modern CTF benchmark problems in dynamical systems.  There is a progression of complexity, from static data, to interactive data where it is possible to run control experiments, to living cyberphysical systems. Top row: (left) Imagenet dataset~\cite{deng2009imagenet}; (middle) Atari video game environment in Gymnasium~\cite{ravichandiran2018hands,dutta2018reinforcement}; (right) pendulum on a cart used for PILCO learner~\cite{deisenroth2011pilco}  Bottom row: (left) Fluid flow past a cylinder and lava lamp; (middle) HydroGym cavity flow environment~\cite{HydroGym}; (right) Double pendulum~\cite{kaheman2023experimental} and Quanser Aero 2 platform~\cite{traver2022digital}.}
    \label{fig:benchmarks}
\end{figure*}

Importantly, our goal is to provide fair evaluation metrics, not disqualify papers from consideration for publication.  In traditional fields of machine learning, state-of-the-art (SOTA) performance has been perhaps over-emphasized, with algorithms being discounted because of fractions of a percentage point difference in raw prediction score.  
In scientific machine learning, algorithm evaluation is more nuanced: algorithms typically have various strengths and weaknesses, and one goal of the CTF is to provide a diversity of scores to better understand these tradeoffs.   
We wish to promote diverse methodological development, but with rational assessments of performance, something which has yet to be standardized in science and engineering~\cite{mcgreivy2024weak}.  A higher-or-lower score should not be the ultimate goal, rather innovation of methods.  After all, it is difficult for a new concept and algorithm to compete immediately with one which has been developed and deployed over two decades.

\begin{figure}[t]
    \centering
    \includegraphics[width=.5\textwidth]{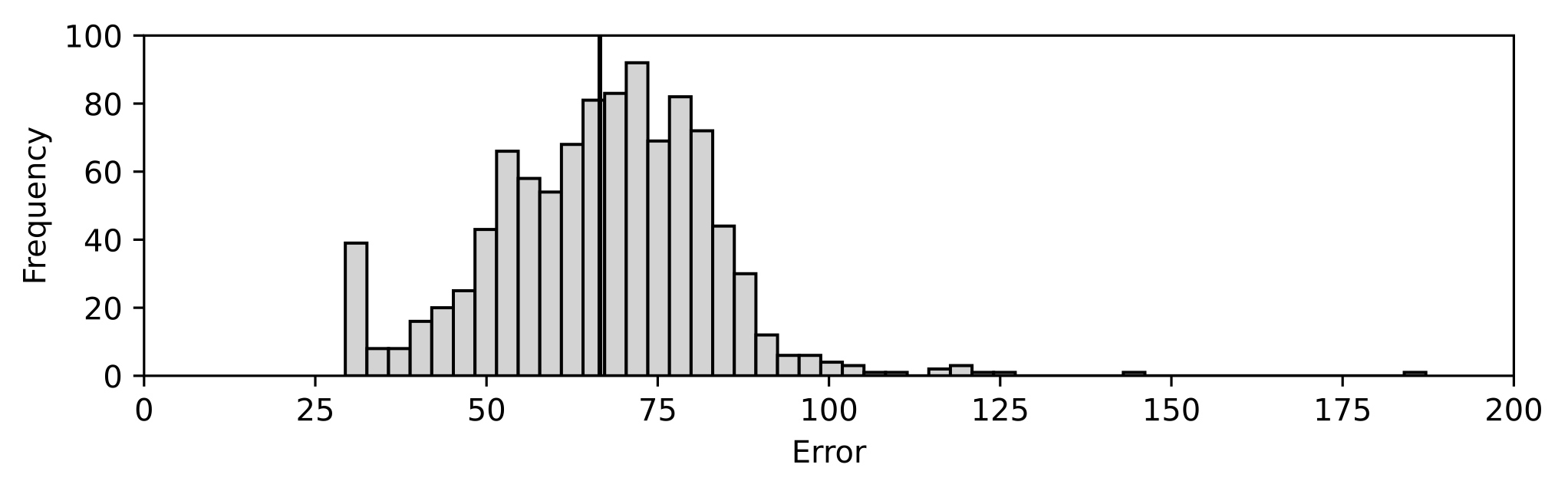}
    \caption{Distribution of errors for fitting the data from 1000 random initial starts of the neural network.  Note that for this application where the test set is in an extrapolation regime, the error distribution is significant, ranging from a minimal RMSE of approximately 29 to a value of 185.  The average error is about 67.  This highlights an important aspect of neural network training:  the performance on a true withheld test set can have very high variance.}
\label{fig:hist_error}
\end{figure}

Although there are currently various benchmarks available to the community (See Fig.~\ref{fig:benchmarks}), they are typically structured as self-reporting tools.  The benchmark environments for dynamic systems, for instance, include those for  ordinary differential equations~\cite{tassa2018deepmind,gilpin2021chaos,dulny2023dynabench,bhamidipaty2024dynadojo} and partial differential equations~\cite{gupta2022towards,ross2022benchmarking,nguyen2023climax,elerevaluating,hao2023pinnacle}, as well as cyberphysical and interactive environments~\cite{hernandez2012decentralized,rajappa2013modelling,traver2022digital,traver2022digitalb,HydroGym,kaheman2023experimental,degrave2022magnetic}.
These benchmarks build on the increasing trend to publish static, though extensive, datasets to test methods, e.g., in fluid mechanics~\cite{perlman2007data,towne2023database}.
However all of these benchmarks involve self testing of algorithms on a known test data set, making it difficult or impossible to have a truly objective comparison.  Moreover, the testing itself is often not consistent with the principle of a withheld test set~\cite{mcgreivy2024weak}.
We argue here that self-reporting is, in general, a flawed premise.  For instance, neural networks upon retraining are typically initialized with random weight assignment.  Although the errors achieved on the training data set are comparable from run to run, the errors on the test set can be significantly different.  This can lead to p-hacking, or judicious picking of results, when reporting scores on test data sets, i.e. simply re-train the model until a desired and good result is achieved for self-reporting.  Only with a true, withheld test set is a comparison among methods possible.  To be precise, consider the simple example of training a feed-forward neural network to fit a simple curve $f(x) = A\cos Bx + Cx + D$ where noise is added to the data to make it more challenging.  The training data is extracted from $x\in[20,40]$ and the model is tested in the domain $x\in[0,50]$.  Thus, the network is required to extrapolate outside of the training regime, which is a typical task in science and engineering.  Each training run provides a high-quality fit (with very little variance from run-to-run) on the training data.  However, when computing the fit on the entire domain, where extrapolation is required, the performance has a large distribution of errors over the various runs as shown in Fig.~\ref{fig:hist_error}.  Selecting the best result from the various training runs may be highly misleading and misrepresentative of the model as a whole.  This would be equivalent to $p$-hacking your results.

\begin{figure}[t]
    \centering
    \includegraphics[width=.5\textwidth]{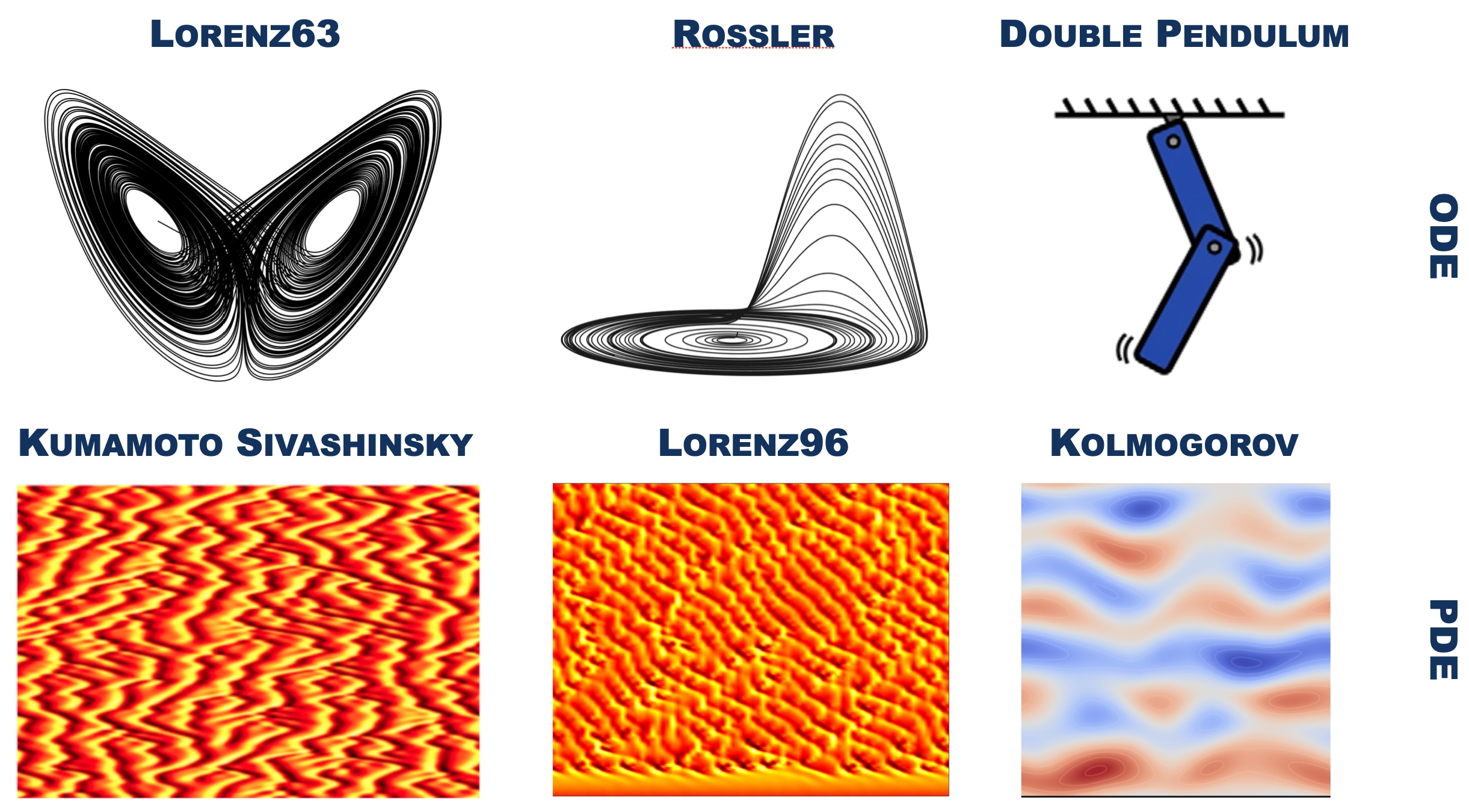}
    \caption{Environments in the AI Institute CTF for dynamical systems permanent collection.  Included in the permanent challenge sets are three dynamical systems (Lorenz, R\"ossler, double-pendulum) and three spatio-temporal systems (Kuramoto-Sivashinsky, Lorenz96, Burgers). The user is provided with 10 training data sets, with the requirement of generating 9 test set approximations.  A diversity of metrics are measured, including those related to forecasting and reconstruction with noisy measurements and limited data.}
\label{fig:ctf_environment}
\end{figure}

\subsection*{Extrapolation versus interpolation}
 
Comparatively speaking, interpolation is easy, extrapolation is hard.   This is for the most part an obvious statement, yet it is often unacknowledged in data science.  The significant successes of machine learning, such as speech recognition, computer vision and autonomy are built upon {\em interpolative} data.   Richard Sutton's blog on the {\em The Bitter Lesson}~\cite{sutton2019bitter} highlights an important theme in machine learning and AI:  simply collect more data for improved models.  In the parlance of curve fitting, collect enough data in order to turn your problem into an interpolation.  Max Welling~\cite{welling2019we} responded to this blog by highlighting that ``The trouble starts when we need to extrapolate.''  Indeed, in the
sciences, the over-arching goal is often aimed at building extrapolatory models capable of building not only understanding, but new technologies.  Extrapolation and interpolation are critical in thinking about scientific models. It suggests that in addition to improving models, we may simply improve by collecting better data.  Both are viable pathways for greater understanding.  However, many science and engineering systems can be difficult to make into interpolation problems, in part because data is often limited.  For instance, in celestial mechanics, once we learn the laws of gravitation, we are capable of envisioning (extrapolating) a moon landing.    Thus for a given data set, one must first determine whether extrapolation, interpolation, or a combination is required.   The answer to this question will often determine the choice of solution techniques.  

Generally speaking, deductive techniques of ML/AI are often overfit and do not extrapolate well.  In contrast, inductive techniques are often constructed for exactly the purpose of extrapolation and generalization.  As an example, one can consider the theoretical foundations of electromagnetism and quantum mechanics.  With these two models established in the late 1800s and early 1900s, we were able over decades to extrapolate our way to creating technologies such as smart phones, which are foundationally based upon these two theoretical concepts.
Thus, in physics-based CTFs, generalization is a critical part of the evaluation metric.  The goal is to  incorporate domain knowledge into data-driven algorithms, or imbue machine learning with physics knowledge and constraints, so that significant and accelerated innovations can be achieved which will allow for testable generalization capabilities.  
Testing for extrapolation also ensures that achieving strong CTF scores, unlike the accuracies achieved in speech and vision, will remain difficult.

\subsection*{Permanent CTF Collection}

The permanent CTF collection includes example toy models of dynamic systems (See Fig.~\ref{fig:ctf_environment}) that are commonly used today in the literature for the evaluation and development of machine learning methods in science and engineering.  These lightweight models are often used due to their simplicity, yet difficulty in producing robust forecasts and reconstructions with noisy measurements and limited data.  The permanent collection is critical as a testbed for method development and fair comparisons.  It is often the case today that vague and misleading statements are made about the success of one method and the failures of another without any fair and comparative testing.  The permanent CTF helps to bring transparency and rigor to emerging methods, helping suppress misleading claims and level the playing field for all.  The primary goal of the permanent CTF collection is to become an essential part of any manuscript:  providing a rigorous evaluation of a given method's capabilities across a diversity of objectives.  This would be critical for fostering the trust and credibility of an algorithm.  With the rapid increase of machine learning methods development for science and engineering, it is important that leading approaches be benchmarked on diverse metrics, including such prominent methods as physics informed neural networks (PINNs)~\cite{Raissi2019jcp,failure21_TR}, operator networks~\cite{lu2021learning,li2020fourier,pan2023neural}, sparse model discovery~\cite{Brunton2016pnas,Rudy2017sciadv} and symbolic regression~\cite{cranmer2023interpretable}, Hamiltonian~\cite{greydanus2019hamiltonian} and Lagrangian~\cite{cranmer2020lagrangian} neural networks, graph neural networks~\cite{battaglia2018relational,sanchez2020learning}, reservoir computing~\cite{pathak2018model}, long-short-term memory networks~\cite{vlachas2018data}, linear dynamic mode decomposition (DMD)~\cite{Schmid2010jfm,Kutz2016book,schmid2022dynamic} and Koopman theory~\cite{Mezic2005nd,Mezic2013arfm,Brunton2022siamreview}, and operator inference~\cite{qian2020lift}, to name only a few.

\subsection*{Rotating CTF Collection}


In addition to a testing platform for method development and evaluation, the CTF will feature a rotating collection of challenging real-world data sets selected broadly from across science and engineering disciplines.  These will be in disciplines as diverse as smart buildings, robotics, brain-machine interfaces, flow control, etc.   Data sets and appropriate domain goals and metrics will be solicited that best fit with the CTF framework.   By providing data and clear goals, a broad group of participants can then play a role in attempting to develop algorithms for accomplishing the tasks assigned.  For each data set chosen, a one-page summary of the data set will be given by its curators.  The summary will outline the importance of the problem and the aspects of the problem that make it unique and challenging.  This allows broad participation since one does not have to be a domain expert or collect data in order to try to advance a given field.  Moreover, the evolution of the CTF will aim to incorporate full cyber-physical systems into the CTF infrastructure in order to test methods on real systems for data-driven control and estimation.  Such an effort can greatly help accelerate the advancement of methods for implementation in real design systems.

\subsection*{The Referee:  Sage Bionetworks}
Sage Bionetworks (sagebionetworks.org) has emerged as a leading platform for enabling the rapid acceleration in biomedical discoveries and the transformation of medicine.  The Sage Bionetworks platform is a flexible framework ideally suited for the sciences and engineering needs of the CTF.   Specifically, Sage Bionetworks offers an easy-to-use framework whereby solutions of the CTF can be uploaded and tested against the withheld test set.  They are the only ones with access to the test set in order to ensure fair comparisons and rigorous evaluations.  A scoreboard is kept for the various tasks assigned for each challenge. Each team competing will be required to share a GitHub link for reproducing the results on the scoreboard.  This is in keeping with the highly successful strategies used in vision, speech and language processing.

\section*{Historical Context}
The computational linguist Liberman coined the notion of the CTF as the set of publicly available data, an agreed-upon scoring function that judges the performance, and competing methods that are being evaluated on the data set based on the scoring function.  Donoho outlines these critical components:  (i) a publicly available training dataset involving, for each observation, a list of (possibly many) feature measurements, and a class label for that observation, (ii) a set of enrolled competitors whose common task is to infer a class prediction rule from the training data, and (iii) a scoring referee, to which competitors can submit their prediction rule.  Importantly, the referee runs the prediction rule against a sequestered testing dataset which is not accessible except by the evaluation algorithm. The referee objectively and automatically reports the score (prediction accuracy) achieved by the submitted rule.   Thus competitors share the {\em common task of training a prediction rule} which they hope will achieve a good score.  The \$1M Netflix 
Challenge~\cite{bennett2007netflix,bell2007lessons}, where the common task was to predict Netflix user movie selections, is perhaps the earliest and most famous of the CTFs.  Kaggle, for instance, has built upon this CTF framework and now offers more than 500,000 public data-sets and 400,000 open source notebooks which users can download and explore winning solutions.  Within such a diversity of data and solutions, it is expected that many new data sets can be explored with previously vetted algorithmic techniques.   It should be noted that CTFs are tools to judge empirical rigor but are, by design, incapable of judging mathematical rigor. Thus, a CTF can only paint an incomplete picture of a methods merit.

\begin{figure}[t]
     \centering
     \includegraphics[width=\linewidth]{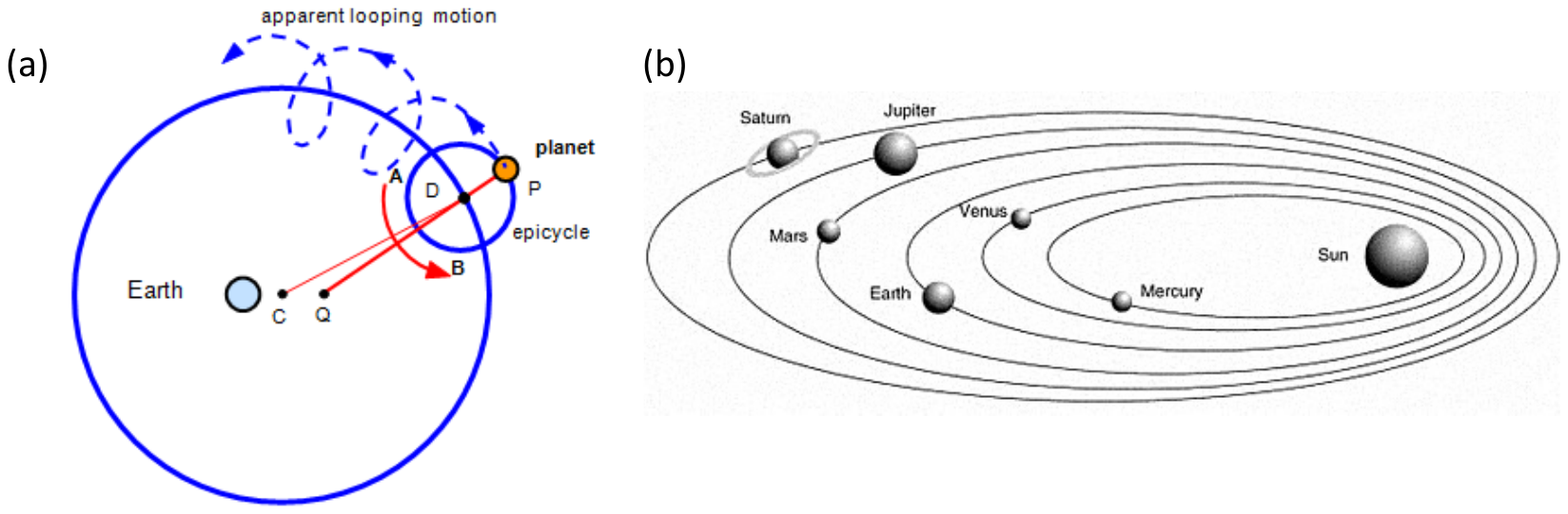}\\[-1.1in]
     \includegraphics[width=\linewidth]{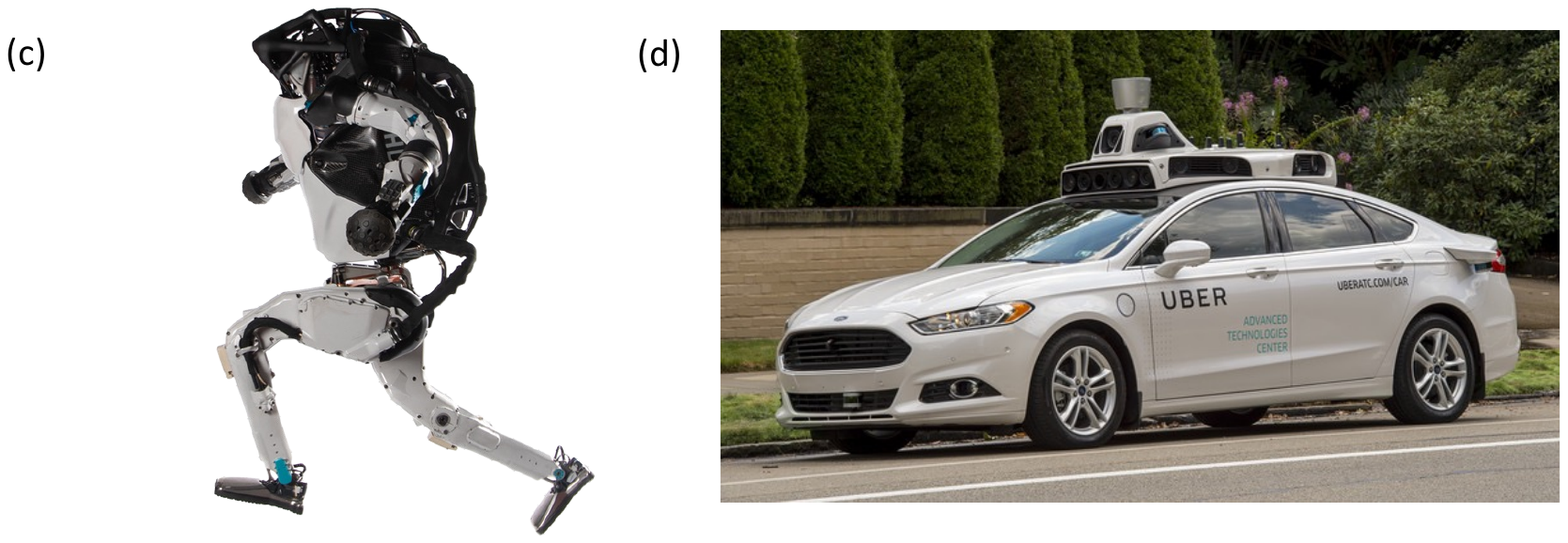}
	\vspace*{-1.2in}
     \caption{(a) From the 2nd century AD until the beginning of the scientific revolution (circa 1600), the {\em deductive} Ptolemaic system of the solar system was used to describe the motion of the planets as {\em circles on circles}.  One can consider this as the precursor to the Fourier transform.  (b)  The inevitable push to a heliocentric coordinate system by Copernicus, Kepler and Galileo allowed Newton to develop an {\em inductive} theory of the solar system:  ${\bf F}=m{\bf a}$. In modern data-driven science, deep learning can be framed towards (c) {\em inductive} models using autoencoder strategies to reduce dynamics to their intrinsic dimensions, or (d) {\em deductive} models which project to higher dimensional spaces and aim for accuracy and flexibility.   Modern autonomy leverages sensors to build data-driven models which are physics-based {\em inductive} models for robotics (e) and which are non-physics based, deductive statistical models for self-driving cars.
     \label{fig:pnas_challenge}}
 \end{figure}

Given the long-standing computational sophistication of the engineering and physical sciences, especially in regards to {\em high-performance computing} (HPC) simulations of complex, multi-scale and multi-physics systems, it is somewhat surprising that the scientific community has not adopted, or perhaps even initially proposed, the CTF architecture.  In part, we venture to suggest that the slow adoption in the natural sciences of the CTF is largely due to the predominance of the {\em inductive} reasoning in this community.  Specifically, physics-based models traditionally are posited from empirical observations and first-principles derivations, which include the consequences of conservation laws,  physical constraints, self-consistent qualitative models, and expert knowledge.  Thus {\em interpretable} governing equations are the starting point for computationally-oriented scientific studies.  In fact, one does not compute until a reasonable set of governing equations are developed.   Validation of these models with data is often a painstaking and time consuming process, yet provides a critical assessment of the inductive reasoning process.  
In contrast, the CTF is a manifestation of the {\em deductive} reasoning process.  Fundamentally, a CTF creates a rigorous and fair framework to evaluate and compare theories or methods from a predominately deductive perspective. Note that unlike inductive approaches that usually try to assign a Boolean truth value that indicates whether or not the conclusion can be derived from the axioms, CTFs mostly assign a soft (or fuzzy)  real-valued score to a method. Liberman argues that improvements to this score often happen at a steady percentage rate until an asymptote is reached which depends on the data quality.
In modern times, induction and deduction reasoning paradigms are both highly successful (see Fig.~\ref{fig:pnas_challenge}), driving innovations in autonomy technologies such as robotics (primarily inductive, physics-based) and self-driving
cars (primarily deductive, sensor-based).

Much like computer vision and speech recognition from a decade ago, the natural sciences are on the brink of a revolution that will be driven by the emergence of high-quality data sets across disciplines that are amenable to the CTF architecture.  Importantly, the CTFs will allow for true comparatives between computing and modeling strategies for diverse scientific efforts, something that has been severely lacking in most fields of application today.  Accountability is critical in this emerging data-driven space, as methods are being developed at an exceptionally rapid pace without proper evaluation~\cite{mcgreivy2024weak}.  In this paper, we aim to provide a platform for researchers in the natural sciences to propose CTFs and compare their methods across a diverse set of scientific tasks such as forecasting, state reconstruction, and control. In the next section, we further the argument for more deductive rigor in the natural sciences, and in section 3, we discuss the requirements of such a platform and what constitutes `good' CTFs. We then introduce example common task frameworks that we believe to have a positive and meaningful impact.

\section*{Physics-Based Models:  Induction versus Deduction} 

Since antiquity, there has been a debate about the nature of scientific discovery. While Aristotle argued that scientific discovery should emerge from observation, his mentor Plato was of the opinion that observation is inherently imperfect and that reason should be the foundation of scientific inquiry. Plato and Aristotle laid the groundwork for the debate about deductive versus inductive reasoning.  The era of enlightenment saw the first substantial evolution of these ideas.  Rationalists like Descartes believed in the ``rational structure'' of reality and that every sensory experience could be a mere illusion and should therefore be doubted.  Empiricists, typified by Hume and Locke, argued that scientific knowledge can only be \emph{a posteriori}, i.e., every scientific insight comes from experience and that before experience the human mind is a blank slate (\emph{tabula rasa}).

Fundamentally, the debate about inductive and deductive reasoning is a debate about the value of scientific discoveries. Deductive and inductive reasoning judge different aspects of a theory. Deductive reasoning tries to impose \emph{empirical rigor}, i.e., it primarily judges a scientific theory by how well it can explain reality, whereas inductive reasoning tries to impose \emph{mathematical rigor}, i.e., it primarily judges a scientific theory by how well it can be explained in terms of axioms of the respective discipline.
More recently, Leo Breiman highlighted much of the modern and applied differences in inductive versus deductive reasoning with his analysis of the two cultures of statistical models~\cite{breiman2001statistical}.  Specifically, he highlighted the difference between {\em statistical learning}, which roughly is inductive in nature, and {\em machine learning}, which leans towards deductive reasoning.  The former aims to provide {\em interpretable}, potentially {\em parsimonious} models, while accuracy and flexibility are the primary goals of the latter.

\subsection*{A case study:  celestial mechanics}

The legacy of the philosophies of Plato and Aristotle pervades the history of celestial mechanics, one of the first and greatest grand challenge problems in science.  The history of celestial mechanics manifests aspects of both inductive and deductive scientific reasoning.  The first successful theory for predicting (forecasting) planetary motion was developed by Claudius Ptolemy in the 2nd century A.D. in Alexandria, Egypt.  Commonly known as the {\em doctrine of the perfect circle}, the retrograde motion of planets was described as the evolution of circles on circles, as shown in  Fig.~\ref{fig:pnas_challenge}(a).   This can be thought of as essentially the earliest version of the Fourier transform, as planetary motion was constructed by summing circular orbits of different frequencies and radii.  This was a deductive model powered by hand calculations and the abacus, which lasted nearly a millennia and a half.   Not until Copernicus, Kepler and Galileo did this theoretical construct fall.  Specifically, a coordinate change from earth-centric to helio-centric orbits allowed the construction of a new deductive theory of planetary motion.  Powered by Tycho Brahe's comprehensive, and order-of-magnitude more accurate, data on planetary motion,  Kepler laid the foundation of modern celestial mechanics by proposing {\em Kepler's laws} and the elliptical orbits of the planets.  Within one hundred years of Kepler's theoretical construct, the first inductive theory, what we would now call a generative model, of planetary motion was proposed by Sir Isaac Newton in his {\em Principia}.  Specifically, the relationship between force and acceleration ${\bf F}=m{\bf a}$.  As even more accurate data became available in the 1800s, discrepancies between theory and astronomical observations were noted.  This led directly to Einstein proposing {\em general relativity} in the early part of the 20th century, which was a revised, inductive theory of gravitation.

The history of gravitation shows that both inductive and deductive reasoning played foundational roles in different periods of the development of celestial mechanics.   Before the innovation of calculus, the theoretical constructs of Ptolemy and Kepler were limited in their mathematical expressibility.  This did nothing to diminish the accuracy of their models.  Indeed, despite the success of Kepler's models, it took many decades before it relegated the doctrine of the perfect circle to history.   Newton's model similarly took a great deal of time to displace Kepler's perfectly reasonable model, i.e. ${\bf F}=m{\bf a}$ did not substantially improve accuracy for some time.  What ultimately allowed Newton to revolutionize the world was the ability of the newly developed calculus and ${\bf F}=m{\bf a}$ to extrapolate.  Indeed, such a theory allowed humans the ability to put astronauts in space and land on the moon.  With the corrections of general relativity, the trajectories of satellites can be planned for deep space missions of the modern era.  And even more recently, the gravitational waves predicted by Einstein's inductive theory were only validated by the LIGO experiment in 2015, nearly 100 years after they were first proposed.  Such scientific predictions represent the highest value of inductive reasoning.

\subsection*{The machine learning shift to more deductivism}
 
In some areas there seems to be considerable opposition to deduction.  This is perhaps not surprising, given our standard undergraduate and graduate discipline training in the sciences which revolves around dominant paradigms of inductive models, including well-established theories of fluid dynamics (Navier-Stokes equation), quantum mechanics (Schr\"odinger equation), electrodynamics (Maxwell's equations), mechanics (${\bf F}=m{\bf a}$), etc. Some disciplines consider deductive approaches not to be `real science' because they often do not result in a formal description of a set of phenomena, i.e. governing equations.  There is also concern about how far a radical deductivism approach can advance science~\cite{pearl2014deductive}.  This opinion is held in spite of a series of negative results in formal systems in the early 20th century.  G\"odels incompleteness theorem showed that there are true but unprovable statements in formal systems describing arithmetic on natural numbers. This was followed by Tarskis indefinability theorem which postulates that formal systems are incapable of representing their own semantics. These results are connected to other negative results in computer science such as Hilbert's Entscheidungsproblem and the Halteproblem and they show that inductive descriptions of reality are fundamentally limited. The reason why, despite these results, inductive approaches to scientific discovery are preferable to some might be that induction often results in a deeper understanding of the subject matter, i.e., generalizability and interpretability. Induction is not only concerned with \emph{how} but also \emph{why} things are the way they are. Answering the \emph{why}-question is often significantly more difficult than answering just \emph{how}. But do we always need this level of understanding? Arguably, most native speakers of the English language have very little inductive knowledge of it. Most native speakers do not know why a sentence is strung the way it is but are perfectly able to generate and understand English language sentences. Thus, for some applications, a deductive understanding of the subject matter suffices. In a sense, when the goal is to produce accurate descriptions of the world, induction can be a detour. Induction relies on a formal description first from which predictions follow, whereas deductive approaches often allow for immediate predictions.
 
 Furthermore, we believe that overly focusing on mathematical rigor or a logical argument can impede or, when done in bad faith, even corrupt the scientific process. One of the most successful and widely used concepts in engineering are Fourier series and the associated Fourier transform. Arguably, the discovery of Fourier series led to a revolution in numerous fields.   But when Joseph Fourier introduced the concepts, a panel responded with: ``... the manner in which the author arrives at these equations is not exempt of difficulties and that his analysis to integrate them still leaves something to be desired on the score of generality and even rigour.'' Furthermore, the authors of~\cite{lipton2019troubling,wornow2023shaky} discuss what they perceive as ``Troubling Trends in Machine Learning Scholarship.'' One pattern they identify as problematic is what they call ``mathiness.'' Specifically, they argue that in order to overwhelm or impress reviewers, scientists might create the illusion of technical depth by including unnecessary or ``\emph{spurious theorems} [...] to lend authoritativeness.'' Thus, inductive reasoning is not free of impurity and overemphasizing inductive merit can lead to the undervaluing of some and to overvaluing of other scientific contributions.
 
 Note that we are not arguing that induction is fundamentally inferior to deduction. As described earlier, we believe that inductive and deductive reasoning judge different aspects of a scientific contribution. They are complementary and ideally every scientific contribution is mathematically sound and empirically accurate. What can be measured by a CTF is fundamentally limited. By design, a CTF is incapable of judging inductive rigor. A CTF can only paint an incomplete picture of a methods merit. On top of that, when we call for a shift to greater empirical rigor in the natural sciences, we are not arguing that the insights gained from the current period of `normal science' should be forgotten. On the contrary, we believe that incorporating domain knowledge into data-driven algorithms should improve their performance. We conjecture that fields in which domain-knowledge can easily be incorporated into data-driven algorithms will outpace fields for which this is harder.  Said another way, dovetailing inductive approaches into deductive approaches, i.e., imbuing machine learning with physics knowledge and contraints, can lead to significant innovations.

 \begin{figure*}[t]
    \hspace*{-.25in}
    \includegraphics[width=1.05\textwidth]{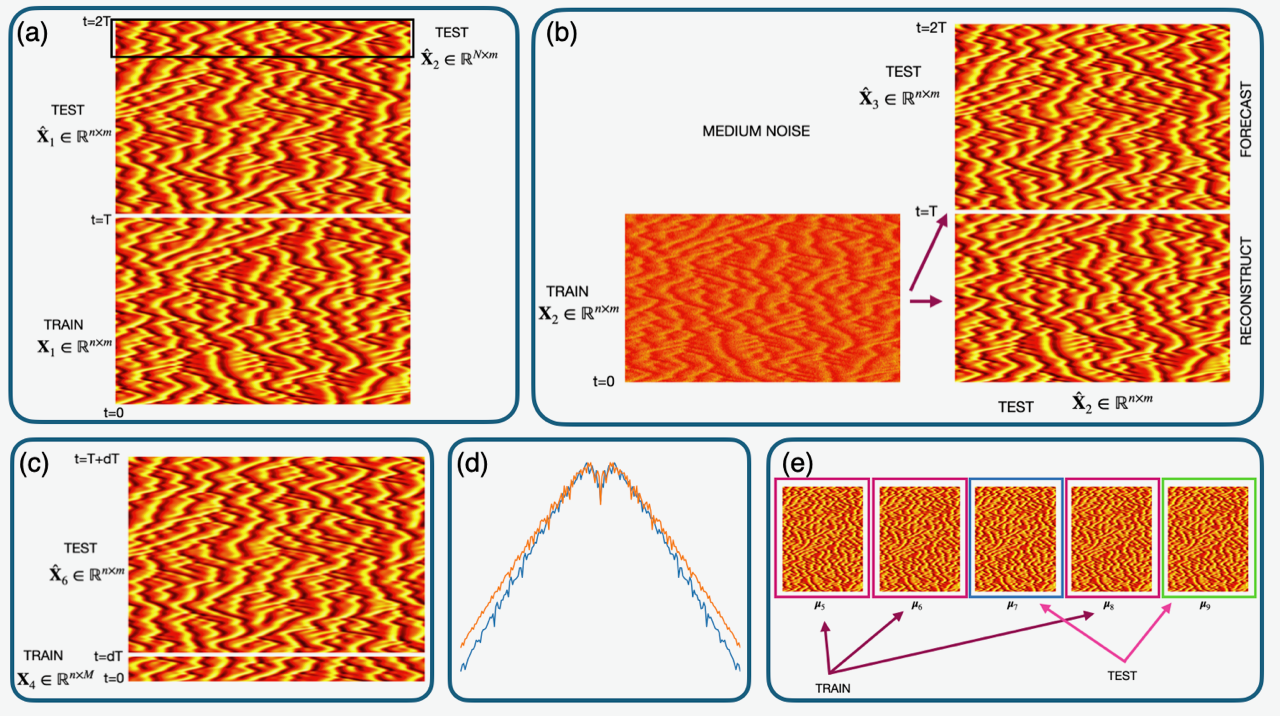}
    \caption{Evaluation of forecasting and reconstruction  capabilities of a method with numerically accurate data for both (a) short- and (b) long-time forecasting, (c) with limited data, (d) with and without noise,  and (e) under parametric variability.}
    \label{fig:ctf}
\end{figure*}

 \section*{Requirements of CTF platforms}
 
Fundamentally, CTFs are a tool to measure empirical progress of a research community in the broad understanding of a specific problem. Considering that improvements to this score seem to happen at a steady percentage rate, CTFs at some point ``get solved.'' To give an example of this, in the past, the MNIST dataset was considered a standard benchmark for image classification algorithms. However, because even fairly simple algorithms exhibit superhuman performance on MNIST, more challenging datasets such as Imagenet and COCO soon emerged. Thus, CTFs are fundamentally temporary and reflect the current challenges researchers in a field face. Because of this, we not only aim to provide high quality datasets and metrics but also a platform for CTFs to live on, evolve and eventually get cataloged because they are considered solved.
 
The {\em common task framework} (CTF) for dynamic systems aims to evaluate algorithms and methods on a variety of tasks that are common for engineering and science.  The goals include forecasting and reconstruction of time-series and spatio-temporal data under the challenges of limited data, noise and parametric dependence.

What will be provided to challengers is a compressed file which includes 11 matrices
\begin{equation}
 {\bf X}_{j} \in \mathbb{R}^{n_j \times m_j} 
 \,\,\,\,\,\,     j=1,2,\cdots, 11
\end{equation}
where
\begin{eqnarray*}
    && n_j = \, \mbox{dimension of dynamical system} \\
    && m_j = \, \mbox{number of time points}\\
    && j = \, \mbox{parameter regime for the} \,j\mbox{th matrix} .
\end{eqnarray*}
Thus, the rows represent the dimension of the system under consideration and the columns are the sequential temporal sampling of the dynamics.  

Twelve tasks and evaluations are assigned to the challengers, with a matrix produced for each one.  Each matrix will be of the following form
\begin{equation}
 \hat{\bf X}_{Jtest} \in \mathbb{R}^{n_J\times m_J} 
 \,\,\,\,\,\,     J=1,2,\cdots, 12
\end{equation}
where $n_J$ (dimension or dynamics) and $m_J$ (number of time points) will be specified for each of the 12 evaluations.  The users approximation to the test set will be ${\bf X}_{Jtest}$ so that ${\bf X}_{Jtruth}$ and ${\bf X}_{Jtest}$ will be compared in the evaluation metrics.\\

\noindent {\bf SCORING}\\

\noindent
Scoring will be on a scale, with 100 being a perfect score and a score of zero corresponding to a guess of zeros for ${\bf X}_{Jtest}$.  Negative scores will show that the model is worse than guessing zeros.  In summary, we have
\begin{eqnarray*}
&& \hspace*{-.2in}  \mbox{SCORE} = 100: \,\,\,  \mbox{Perfect match between model and truth}   \\
&&   \hspace*{-.2in} \mbox{SCORE} = 0: \,\,\,  \mbox{Score for model with zeros}
\end{eqnarray*}

Thus, the mechanics of {\em each} challenge, three dynamical systems (Lorenz, R\"ossler, double-pendulum) and three spatio-temporal systems (Kuramoto-Sivashinsky, Lorenz96, Burgers), will provide the user 10 training data sets, with the requirement of 9 test set approximations to be returned for evaluation on 12 metrics.  These are detailed in an example using the Kuramoto-Sivashinsky model.

\section*{Example System:  Kuramoto-Sivashinsky}

The Kuramoto-Sivashinsky (KS) equation is a fourth order, nonlinear partial differential equation.  It is considered a canonical example of spatio-temporal chaos  in a one-dimensional PDE, and it is therefore commonly used as a test problem for data-driven algorithms.  The KS equation is a particularly challenging case for fitting algorithms due to its combination of high dimensionality, nonlinearity, and sensitivity to initial conditions (chaotic behavior):
\begin{equation}
u_t+uu_x+u_{xx}+\mu u_{xxxx} = 0 \label{eq:KS} .
\end{equation}
Solutions of the (\ref{eq:KS}) are on a grid $[0, 32\pi]$ with periodic boundary conditions.  A numerical integrator with unknown time-step $\Delta t$ will be used to evolve the solution forward $m$-steps.  

\subsection{Test 1:  Forecasting (2 scores)}

The first test of the method, as illustrated in Fig.~\ref{fig:ctf}(a), involves the approximation of the future state of the system.  Thus, given a data matrix representing the dynamics from $t\in[0,T]$ (${\bf X}_{1}\in \mathbb{R}^{n_1 \times m_1} $), the forecast requested is from $t\in[T,2T]$ ($\tilde{\bf X}_{1}\in \mathbb{R}^{n_1 \times m_1} $).  The forecasting score is actually composed of two scores evaluating both the short-time forecast (the ``weather forecast'') which is computed using root-mean square error fitting between the test set and the users approximation, and the long-term forecast (the ``climate forecast''), which is based upon the power spectral density.  As such, the following two error scores are computed:
\begin{equation}
   E_{{\mbox{\tiny  ST}}}=\frac{\| \hat{\bf X}_1 [:,1:k] - \tilde{\bf X}_1 [:,1:k]  \|}{\|\hat{\bf X}_1 [:,1:k]\|}
   \hspace{.2in} \mbox{(weather forecast)}
   \label{eq:st}
\end{equation}
and
\begin{equation}
    E_{{\mbox{\tiny  LT}}}\!=\!\frac{\| \hat{\bf P}_1 [{\bf k},N\!-\!k\!:\!N] \!-\! \tilde{\bf P}_1 [{\bf k},N\!-\!k \!:\!N] \|}{\| \hat{\bf P}_1 [{\bf k},N\!-\!k:N]\|}
     \hspace{.1in} \mbox{(climate forecast)}.
\end{equation}
Here, $E_{ST}$ is the short-time error evaluated using least-squares and $E_{LT}$ (See Fig.~\ref{fig:ctf}(b)) is the long-time error which is computed by least-squares fitting of the power spectrum ${\bf P}_j [:,k] =\ln ( |\mbox{FFT}({{\bf X}_j[:,k]})|^2 )$, where the {\bf fftshift} has been used to model the data in the wavenumber domain and ${\bf k}=n/2-k_{m}:n/2+(k_m+1)$ with $k_m=100$. This then looks at the first 100 wavenumbers in order to determine the decay rate of the power spectrum.
It is clear that there are many ways to evaluate the long-range forecasting capabilities.  However, we have chosen a simple metric, fully understanding that more nuanced scoring could be used.

The first error metric evaluates the overall forecasting skill for short-time prediction:
\begin{equation}
    E_1= 100 (1 - E_{{\mbox{\tiny  ST}}} ).
    \label{eq:error1}
\end{equation}
Note that as a baseline a solution guess of zeros $\tilde{\bf X}_1 [:,1:k]={\bf 0}$ gives a score of $E_1=0$.
The second error metric evaluates the long-range capabilities of an algorithm for matching the correct long-term behavior of the system
\begin{equation}
    E_2= 100 (1 - E_{{\mbox{\tiny LT}}}).
\end{equation}
Note that as a baseline a solution guess of zeros $ \tilde{\bf P}_1 [{\bf k},N-k:N]={\bf 0}$ gives a score of $E_2=0$.
For the challenge dynamics of interest, sensitivity of initial conditions is common so that long range forecasting to match the test set is not a reasonable task given fundamental mathematical limitations with Lyapunov times.\\

{\bf Input:} ${\bf X}_{1train} $

{\bf Output:} ${\bf X}_{1test}$

 {\bf Scores:} $E_1, E_2$

\begin{figure}[t]
    \centering
\includegraphics[width=0.45\textwidth]{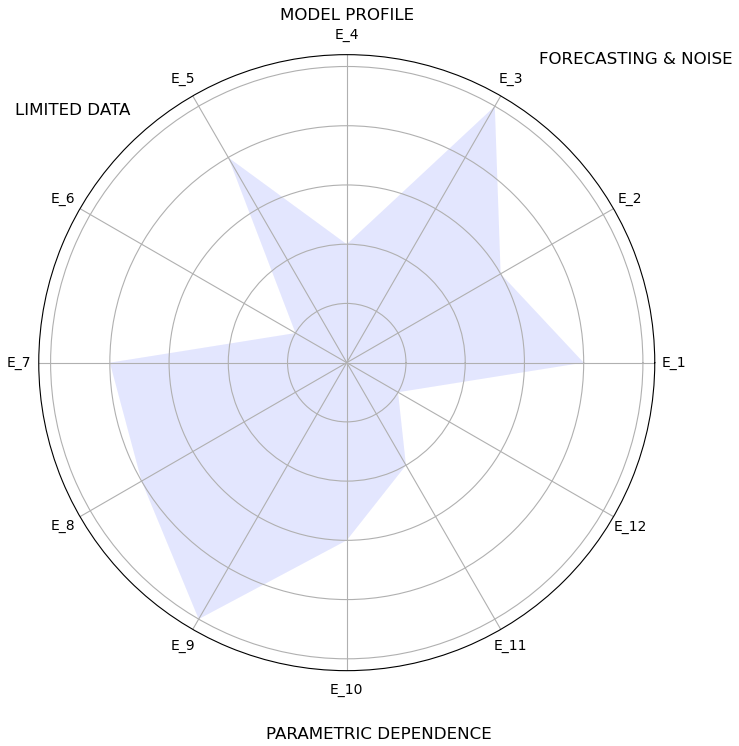}
    \caption{Evaluation of a method's capabilities across the twelve evaluation metrics.  The radar plot profiles how well a method does on the various tasks associated with forecasting and reconstruction with noise, limited data and parametric dependency.}
    \label{fig:radar}
\end{figure}

\subsection{Test 2: Noisy Data (4 scores)}

The ability to handle noise is critical in all data-driven applications, as sensors and measurement technologies are by default embedded with varying levels of noise.  Methods that work with numerically accurate data, for example data points that are $10^{-6}$ accurate, may be useful for model reduction, but they are rarely suitable for discovery and engineering design from real-world data.  Both strong and weak noise are considered as these represent realistic challenges to be addressed in practice.

Figures~\ref{fig:ctf}(d)  demonstrates the challenges to be addressed.  The challenge is very similar to Test 1, but now with noise added to the data. Specifically, what is given to the challenger is a data matrix   ${\bf X}_{2train} $ and ${\bf X}_{3train}$ representing the evolution over with medium or high noise.  The objective is to first produce a reconstruction of the data itself, i.e., denoise the data to produce an estimate of the true state of the dynamics, and the second objective is to then forecast the future state, matrices ${\bf X}_{2test} $ and ${\bf X}_{3test} $ for ${\bf X}_{2train}$ and matrices ${\bf X}_{4test} $ and ${\bf X}_{5test}$ for ${\bf X}_{3train}$.  For the first task, a least-square fit is used between the approximation of the denoised data and the truth, which is given by $E_1$ for all snapshots $m$. The forecasting score is given by the error metric $E_2$.  Thus, two error scores $E_3$ (reconstruction) and $E_4$ (forecast) are produced for medium noise, and two error scores $E_5$ (reconstruction) and $E_6$ (forecast) are produced for high noise.\\

{\bf Input:} ${\bf X}_{2train}$, ${\bf X}_{3train} $

{\bf Output:} ${\bf X}_{2test}$, ${\bf X}_{3test} $, ${\bf X}_{4test} $, ${\bf X}_{5test} $

 {\bf Scores:} $E_3, E_4, E_5, E_6$

\subsection{Test 3: Limited Data (2 scores)}

Data limitations are also present in many systems, which often change which data-driven architectures are most successful.  The low-data limit is critically important in many applications in engineering and science, thus requiring the evaluation of methods under these conditions.

Figure~\ref{fig:ctf}(c) demonstrates the nature of the test.  In this case, only a limited number of snapshots $M$ on numerically accurate data are given ${\bf X}_{4test}\in \mathbb{R}^{n \times M} $.  From this limited data, a forecast must be made which is evaluated with the error metrics both $E_1$ and $E_2$ on the approximated future  ${\bf X}_{6test} $.  The experiment is repeated with noise on the measurements using the training matrix  ${\bf X}_{5train} $ for which a forecasting prediction matrix is produced 
${\bf X}_{7test}\in \mathbb{R} $.  Two error scores ($E_1$ and $E_2$) are produced for the noise-free and noisy limited data.  These scores are $E_7$ (short) and $E_8$ (long) for the noise free case and $E_9$ (short) and $E_{10}$ (long) for the noisy case.\\

{\bf Input:} ${\bf X}_{4train} $, ${\bf X}_{5train} $

{\bf Output:} ${\bf X}_{6test} $, ${\bf X}_{7test} $

 {\bf Scores:} $E_7, E_8, E_9, E_{10}$

\subsection{Test 4:  Parametric Generalization (4 scores)}

Finally, the ability of a model to generalize to different parameter values is evaluated.  For this case, the model's ability to interpolate and extrapolate to new parameter regimes is considered with noise-free data and noisy data as well.  The interpolation and extrapolation are each their own score.  This gives a total of four scores that evaluate parametric dependence.

Figure~\ref{fig:ctf}(e) shows the basic architecture of the test.  For the noise-free case, three training data sets are provided with three different parameter values ${\bf X}_{6train} $, ${\bf X}_{7train} $ and ${\bf X}_{8train} $. Construction of the dynamics in parametric regimes that are interpolatory ${\bf X}_{8test} $ and extrapolatory ${\bf X}_{9test}$ are required.  for both of the test regimes, a burn in matrix is given .  The error metric $E_1$ is used to evaluate the reconstructions of the interpolatory and extrapolatory regimes ${\bf X}_{9train}$ and ${\bf X}_{10train} $ respectively.\\

 {\bf Input:}  \!${\bf X}_{6train}$, ${\bf X}_{7train} $, ${\bf X}_{8train} $,  
 
 \hspace*{.45in}${\bf X}_{9train} $, ${\bf X}_{10train} $

{\bf Output:} ${\bf X}_{8test}$, ${\bf X}_{9test}$

 {\bf Scores:} $E_{11}, E_{12}$

\subsection*{Summary Evaluation}

To evaluate the overall performance of a method, a radar plot is developed highlighting the various scores associated with the challenge.  Figure~\ref{fig:radar} shows how each method will look overall.  It is a profile of the method rather than a single score.  Of course, the average of all scores can be computed in order to provide a composite score.  But ultimately, different tasks will excel in different areas.  Some will do well with noise, others will not.  Others might excel in the limited data regime, while being poor under parametric generalization.  Profiles are important in order to provide a more comprehensive and well-rounded metric of performance.

\section*{Outlook and Discussion}

In summary, it is time for the engineering sciences to have a stable, robust and rigorous CTF, both for promoting accountability and for accelerating advancement of machine learning methods.  More than just benchmarking, the CTF aims to provide quantifiable metrics that are rigorously evaluated across a diverse set of tasks.  It is necessary that the community has a fair assessment of the multitude of methods being developed.  The advancement of machine learning and AI for science and engineering will rely heavily on fair evaluations and open-source code, all of which are directly built in to the CTF framework.

And the CTF is open to everyone!  We are encouraging practioners across the sciences and engineering to participate in the fair assessment of their algorithms in order to promote a culture of accountability and fairness.  The CTF examples can all be evaluated with laptop level computing, and as a result does not have a barrier to entry.  Thus graduate students across institutions can easily participate in the development of algorithms.  This is intentional as many modern machine learning algorithms now require computational platforms and data resources well beyond the capabilities of most academic research groups.

\section*{Acknowledgments}
The authors acknowledge support from the National Science Foundation AI Institute in Dynamic Systems (grant number 2112085). SLB acknowledges funding support from The Boeing Company.


\bibliography{pnas-sample}

\end{document}